# Local reservoir model for choice-based learning

**Makoto Naruse[1*], Eiji Yamamoto[2], Takashi Nakao[3], Takuma Akimoto[4], Hayato Saigo[5], Kazuya Okamura[6], Izumi Ojima[7], Georg Northoff[8**], and Hirokazu Hori[9]**

[1] Network System Research Institute, National Institute of Information and Communications Technology, Koganei, Tokyo, Japan

[2] Graduate School of Science and Technology, Keio University, Yokohama, Kanagawa, Japan

[3] Department of Psychology, Graduate School of Education, Hiroshima University, Hiroshima, Japan

[4] Department of Physics, Faculty of Science and Technology, Tokyo University of Science, Noda, Chiba, Japan

[5] Nagahama Insitute of Bio-Science and Technology, Nagahama, Shiga, Japan

[6] Graduate School of Informatics, Nagoya University, Nagoya, Aichi, Japan

[7] Shimosakamoto, Otsu, Shiga, Japan

[8] Mind, Brain Imaging and Neuroethics Research Unit, The Royal's Institute of Mental Health Research, University of Ottawa, Ottawa, Canada

[9] Interdisciplinary Graduate School, University of Yamanashi, Kofu, Yamanashi, Japan

\* naruse@nict.go.jp

** Equal contribution to last author

## Abstract

Decision making based on behavioral and neural observations of living systems has been extensively studied in brain science, psychology, neuroeconomics, and other disciplines. Decision-making mechanisms have also been experimentally implemented in physical processes, such as single photons and chaotic lasers. The findings of these experiments suggest that there is a certain common basis in describing decision making, regardless of its physical realizations. In this study, we propose a local reservoir model to account for choice-based learning (CBL). CBL describes decision consistency as a phenomenon where making a certain decision increases the possibility of making that same decision again later. This phenomenon has been intensively investigated in neuroscience, psychology, and other related fields. Our proposed model is inspired by the viewpoint that a decision is affected by its local environment, which is referred to as a local reservoir. If the size of the local reservoir is large enough, consecutive decision making will not be affected by previous decisions, thus showing lower degrees of decision consistency in CBL. In contrast, if the size of the local reservoir decreases, a biased distribution occurs within it, which leads to higher degrees of decision consistency in CBL. In this study, an analytical approach for characterizing local reservoirs is presented, as well as several numerical demonstrations.

Furthermore, a physical architecture for CBL based on single photons is discussed, and the effects of local reservoirs is numerically demonstrated. Decision consistency in human decision-making tasks and in recruiting empirical data are evaluated based on local reservoir. In summary, the proposed local reservoir model paves a path toward establishing a foundation for computational mechanisms and the systematic analysis of decision making on different levels (such as from photons to brains). This foundation may lead to further endeavors in future decision sciences and applications.

**Introduction**

Decision making based on behavioral and neural experimental findings has been studied in a variety of disciplines, ranging from neuroscience to neuroeconomics [1-5]. Decision making also forms a foundation for artificial intelligence [6]. For instance, artificially constructed physical decision-making mechanisms have been recently experimentally implemented using single photons [7] and chaotic lasers [8]. Because decision making demands that a choice be made between two or more various alternatives on both the neuronal level of the brain and at the physical level of photons or chaotic lasers, a common foundation—in the form of a specific computational mechanism—must be assumed on different levels, regardless of its physical realizations. This concept is schematically summarized in Fig 1(a).

One commonality in decision making on different levels is that previous decisions may impact current and future decisions. For instance, choosing one option may cause one to choose it again in the future. Such decision consistency has been described as choice-based learning (CBL), or more precisely—choice-induced preference change, which has been extensively studied in neuroscience, psychology, and other related fields [9-14]. However, the underlying computational mechanisms of CBL remain unclear.

In this study, we propose a specific model, the local reservoir model, as one computational mechanism that drives decision consistency in CBL. The local reservoir model highlights the hidden architecture (or environments) behind decision making that naturally incorporates the intrinsic attributes of the entities. In addition, the model accommodates uncertainties or fluctuations in a systematic manner. In the local reservoir model, a decision between two possible choices is represented as an energy dissipation to either of two lower energy states; dissipation to the "left" state is correlated with one decision, while dissipation to the "right" state is associated with the other decision. Importantly, the energy dissipation should be absorbed by the surrounding environments, which are referred to as local reservoirs. If the size of the local reservoir is large enough, consecutive decision making will not be affected by past decisions, thus demonstrating lower degrees of CBL. In contrast, if the size of the local reservoir decreases, a biased distribution could occur in the local reservoir, leading to identical decisions with high degrees of CBL.

This study is organized as follows. First, the local reservoir model is introduced, followed by numerical evaluations in which decision consistency in CBL is clearly observed to be dependent on the size of local reservoir. Then, we present an analytical approach for characterizing the local reservoir and generalizing the dynamic attributes of the model. In addition, a single-photon-based physical architecture for decision consistency in CBL is discussed as a realization of the local reservoir model. Furthermore, empirical data on decision consistency in human decision making are analyzed from the viewpoint of local reservoirs. Finally, we discuss various applications of local reservoir model, such as reinforcement learning [6,15], internally guided and externally guided decision making [10], and self [16] and consciousness [17‑19] at the neural level of the brain.

## Local reservoir model

The local reservoir model is inspired by the category theoretic analysis of decision making [20]. The decision-making issue therein was the multi-armed bandit problem (MAB), in which an accurate and prompt decision is required to choose the most profitable slot machine among many slot machines. The MAB problem is difficult to solve because exploratory action is needed to find the best slot machine, but too much exploration leads to significant losses. At the same time, hasty decision making may result in missing the best machine [6]. The category theoretic study reveals that environmental entities are involved in the decision-making process, and that they appear in the form of an octahedral structure. This means that a total of six entities are interdependent with each other [20,21]. What is important is that a decision is tightly related to the environment, which is schematically summarized as shown in Fig 1(b).

This study was inspired by this realization and examines decision consistency in CBL, or choice-induced preference change, which have been studied in a variety of research [9‑14]. To highlight the most simplified spatial and temporal aspects of a local reservoir and its impact to CBL, we provide discussions using a one-dimensional model. In addition, the local reservoir approach described below can be easily extendable to other learning problems, such as reinforcement learning.

The number of choices considered in this study is two, and are referred to as Decision L and Decision R hereafter. As schematically shown in Fig 2, we consider a diagram that mimics an energy level diagram of quantum nanostructures. In this diagram, one upper energy level (denoted by $A^{(U)}$) and two lower levels ($A_L^{(L)}$ and $A_R^{(L)}$) exist. An elemental excitation (e.g., an electron) can occupy each of these levels while satisfying the condition that the number of excitations sitting on a particular level is only one; namely, excitations are assumed to be fermion. An excitation in the upper level is relaxed to one of the lower levels via energy dissipation. Hereafter, we refer to this diagram as the *visible system*.

We consider the relaxation to the "left" lower level ($A_L^{(L)}$) as Decision L, while the relaxation to the "right" ($A_R^{(L)}$) as Decision R. Importantly, the energy dissipation should be absorbed by the surrounding environments; this is where the local reservoir comes into play. If the surrounding environment cannot accommodate energy dissipation, the excitation in the upper level cannot be relaxed to lower levels; one such phenomena is known as the phonon bottleneck [22].

The model of the reservoir will be formulated next. The number of lower energy levels in the local reservoir is denoted by *N*, which is equal to five in the schematic diagram shown in Fig 2. Each of the energy levels is denoted by $L_i^{(L)}$ where *i* ranges from 1 to *N*. The upper energy levels are labeled by $L_i^{(U)}$, where *i* ranges from 1 to $N+1$. Here, we recollect that Decision L is regarded as a relaxation to the left lower energy level in the apparent system. Correspondingly, one excitation located in the lower energy level in the local reservoir is excited to the upper level on the "right" hand side. For example, the arrow denoted by **L2** in Fig 2 is a rightward excitation from $L_2^{(L)}$ to $L_3^{(U)}$. Actually, the decision (or relaxation) observed in the apparent system stems from such an excitation in the local reservoir. We can equate the decision indicated by the apparent system with an excitation in the local reservoir.

Consequently, because the excitation originally located at $L_2^{(L)}$ is now excited, the possibility of excitation from $L_2^{(L)}$ to $L_2^{(U)}$, marked by **R2**, is zero. In addition, because energy level $L_3^{(U)}$ is now occupied, another excitation resting in $L_3^{(L)}$ cannot be excited to $L_3^{(U)}$, marked by **R3**. As a result, two yellow-colored arrows (**R2** and **R3**) are disabled; energy excitation by these means are unavailable, whereas the disabled green arrow is only **L2**. Therefore, what follows is that the number of green arrows is greater than the number of yellow arrows; hence, an excitation is more likely to be induced via one of the green arrows in the local reservoir in the next step. This means that Decision L is more preferably chosen than Decision R, demonstrating the occurrence of CBL. Specifically, the selection of the first decision (Decision L) leads to a greater likelihood of choosing that same decision (Decision L) in the second round. For the 3$^{rd}$ decision, there are three green arrows and only one yellow arrow in the local reservoir; thus, the likelihood of Decision L is more probable than Decision R.

If the size of the local reservoir is large enough ($N \gg 1$), one decision does not create a significant impact on the local reservoir. The imbalance between the green and yellow arrows is negligible. Therefore, consecutive decisions will not be affected by past decisions, thereby demonstrating lower degrees of CBL. In contrast, if the size of the local reservoir is small, a biased distribution occurs in the local reservoir, as described in the abovementioned example, which leads to a high degree of CBL.

Numerical evaluations were performed to characterize the relation between the properties of local reservoirs and CBL. To avoid to the creation of artifacts after the "initial" state, in which all the upper energy levels of the local reservoir are empty, we evaluated the decisions after sufficient time had elapsed. Meanwhile, as the time elapses, all the upper energy levels could be occupied, meaning that no more decisions are made. In reality, the excitations induced in the upper energy levels are relaxed to *the reservoir of the local reservoir*. Likewise, the empty lower energy levels in the local reservoir are filled with another excitation via its reservoir. We analytically characterize the dynamics of relaxation and excitations associated with the local reservoir in the next section. In the numerical simulations, we introduced the notion of excitation "lifetime" in the local reservoir. When an excitation moves from a lower level to an upper level, three paths are eventually disabled, as described above. We assume that such disabled arrows are recovered after the lifetime-value cycles have elapsed.

More specifically, 3,000 consecutive decisions were made and the sequence was repeated 100,000 times. In evaluating CBL, we first check the decision at the cycle of $T_0 = 2,000$. The *decision consistency* at cycle $T_0 + t$ is defined by one if the decision at $T_0 + t$ is the same as that at cycle $T_0$; otherwise, the decision consistency is given by zero. In the analysis shown in Fig 3(a), the lifetime value was assumed to be 10. The average decision consistency was calculated by 100,000 samples as a function of the cycle $t$ (after $T_0$) with regard to the local reservoir size (*N*) of 4, 10, 20, 30, 40, 50, and 100 as shown in Fig 3(a). As shown in the figure, the decision consistency exhibits a larger value for a smaller-sized local reservoir. At the same time, as the time elapses, for instance when $t$ is larger than 50, the decision consistency converges to 0.5, meaning that there is no correlation between the decisions at $T_0$ and $T_0 + t$. These observations demonstrate that a smaller-sized local reservoir yields a higher degree of CBL.

Figure 3(b) visualizes CBL in the simulation from the perspective of a random walker, where Decision L is evaluated as a unit positional change in the upward direction, whereas Decision R is presented as a unit change in the downward direction. The red-colored tracks depict several walkers beginning with Decision L at cycle $T_0$ whereas the blue-colored tracks represent traces starting with Decision R at cycle $T_0$. With a smaller-sized local reservoir ($N = 4$), the blue traces and red traces are biased upward and downward, respectively, indicating that the chance of conducting the same decisions after cycle $T_0$ increases (Fig 3(b)). However, both the blue and red traces behave similarly to a conventional random walker with a larger-sized local reservoir ($N = 100$) (Fig 3(c)).

Figure 3(d) characterizes the dependence on lifetime value. The maximum decision consistency is evaluated as a function of the size of the local reservoir while the lifetime values are configured in 1, 5, 10, 15, and 20. As the lifetime value increases, the decision consistency exhibits values larger than 0.5, even in larger-sized local reservoirs. This is a natural consequence because a large lifetime value provides a higher degree of upper energy level occupations, or increased appearances of disabled arrows in the local reservoir. It is noteworthy that a lifetime value of 1 yields a decision consistency of around 0.5 *regardless* of the size of the local reservoir. A lifetime value of 1 means that the upper energy levels of the local reservoir are always completely empty, which leads to an equal probability of choosing Decision L and R, no matter the size of local reservoir. Therefore, CBL is not observed, regardless of the size of the local reservoir.

Figure 3(e) evaluated the *active* portion, defined by the percentage of the *used* arrows in the local reservoir at cycle $T_0$, as a function of the size of local reservoir. The active portion decreases as the increase of the size of local reservoir while it increases as the lifetime value increases. This is consistent with the increased decision consistency in the smaller-sized and large-lifetime local reservoir.

## Analytical approach to local reservoir model

The dynamics of a local reservoir are specified by three characteristics: (i) the rate at which the lower energy levels are filled ($\gamma_{in}$), (ii) the rate of excitation from a lower energy level to an upper energy level ($\gamma_{up}$), and (iii) the rate of excitation disappearance from the upper energy levels ($\gamma_{out}$).

For an $N=1$ system (Fig 4(a)) in which there are eight total states concerning the excitation occupation in the upper levels ($L_1^{(U)}$ and $L_2^{(U)}$) and the lower level ($L_1^{(L)}$), let each state be specified by index numbers ($1,\cdots,8$), as shown in Fig 4(a). The states are related to each other by the dynamics of one of the three aforementioned characteristics. For example, the empty state (No. 1) is transferred to the state of owing an excitation in the lower level (No. 2) via excitation fulfilling dynamics ($\gamma_{in}$). Consequently, the rate equation of the local reservoir is given by

$$\begin{pmatrix} dp_1/dt \\ dp_2/dt \\ dp_3/dt \\ dp_4/dt \\ dp_5/dt \\ dp_6/dt \\ dp_7/dt \\ dp_8/dt \end{pmatrix} = \begin{pmatrix} -\gamma_{in} & 0 & \gamma_{out} & \gamma_{out} & 0 & 0 & 0 & 0 \\ \gamma_{in} & -2\gamma_{up} & 0 & 0 & \gamma_{out} & \gamma_{out} & 0 & 0 \\ 0 & \gamma_{up} & -\gamma_{in}-\gamma_{out} & 0 & 0 & 0 & \gamma_{out} & 0 \\ 0 & \gamma_{up} & 0 & -\gamma_{in}-\gamma_{out} & 0 & 0 & \gamma_{out} & 0 \\ 0 & 0 & \gamma_{in} & 0 & -\gamma_{out}-\gamma_{up} & 0 & 0 & \gamma_{out} \\ 0 & 0 & 0 & \gamma_{in} & 0 & -\gamma_{out}-\gamma_{up} & 0 & \gamma_{out} \\ 0 & 0 & 0 & 0 & \gamma_{up} & \gamma_{up} & -\gamma_{in}-2\gamma_{out} & 0 \\ 0 & 0 & 0 & 0 & 0 & 0 & \gamma_{in} & -2\gamma_{out} \end{pmatrix} \begin{pmatrix} p_1 \\ p_2 \\ p_3 \\ p_4 \\ p_5 \\ p_6 \\ p_7 \\ p_8 \end{pmatrix} \quad (1)$$

where $p_i$ ($i=1,\cdots,8$) represents the probability of the occupying state, $i$. In the steady state, $p_i$s is derived by solving Eq (1) and letting the left-hand side be zero, and allowing the condition of the unity of probabilities to be as follows: $\sum p_i = 1$.

Because our interest is focused on trends in consecutive identical decision making, we are concerned with what the probability is for Decision L to be followed by the same decision, referred to as $P(L \to L)$, compared to the probability of Decision L to followed by a different decision (Decision R), denoted by $P(L \to R)$. The *decision transition probability* $P(L \to L)$ consists of a variety of state transitions, such as "$2 \to 4 \to 1 \to 2 \to 4$", "$2 \to 4 \to 6 \to 2 \to 4$". We do not present entire transitions here in order to avoid unnecessarily complex descriptions, while it should be remarked that decision transition probabilities are systematically derived, even as *N* becomes large, and hence the decision transition probabilities are computable. Specifically, the imbalance of decision transition probabilities, $P(L \to L) - P(L \to R)$, is easily and analytically derived when $N = 1$ and is given by

$$\frac{\gamma_{in}\gamma_{up}}{(\gamma_{in}+\gamma_{out})(\gamma_{up}+\gamma_{out})}\left\{\frac{1}{2}(p_1+p_2)+\frac{\gamma_{in}\gamma_{out}+\gamma_{out}(\gamma_{up}+\gamma_{out})}{2(\gamma_{in}+\gamma_{out})(\gamma_{up}+\gamma_{out})}(p_3+p_4)\right.$$
$$\left.+\frac{2(2\gamma_{in}+\gamma_{out})(\gamma_{up}+\gamma_{out})\gamma_{out}^2+\gamma_{in}\gamma_{out}(\gamma_{in}+\gamma_{out})^2}{2(\gamma_{in}+2\gamma_{out})(\gamma_{in}+\gamma_{out})^2(\gamma_{up}+\gamma_{out})}p_7+\frac{\gamma_{out}}{2(\gamma_{up}+\gamma_{out})}(p_5+p_6+p_8)\right\}.$$
(2)

In fact, an $N = 1$ local reservoir does *not* lead to CBL. This can be intuitively understood because the occupation of one of the upper energy levels prohibits the same path of excitations.

An interesting observation occurs in CBL when *N* is larger than two. Because the number of states becomes large (32), the analytical derivation of the explicit forms of state probabilities becomes difficult to perform in practice. However, the procedure is essentially the same as in the above example of $N = 1$; hence, the state transition probabilities are derived in a straightforward manner. We evaluate decision transition probabilities with respect to several representative cases when the relaxation rates $(\gamma_{in}, \gamma_{up}, \gamma_{out})$ are specified by (i) (10,1,1), (ii) (1,10,1), and (iii) (1,1,10) (Fig 4(b)). As characterized in Fig 4(c), the imbalance of decision transition probabilities $P(\text{L} \to \text{L}) - P(\text{L} \to \text{R})$ exhibits a *positive* value with $(\gamma_{\text{in}}, \gamma_{\text{up}}, \gamma_{\text{out}}) = (1,1,10)$, meaning that a consecutive identical decision is much more probable than non-identical decisions. Physically, this condition is consistent with the *larger* lifetime local reservoirs discussed in the previous section; the excitations are clogged within the local reservoir. However, the local reservoir dynamics of $(\gamma_{\text{in}}, \gamma_{\text{up}}, \gamma_{\text{out}}) = (1,10,1)$ indicate that excitations are quickly excited and relaxed to the reservoir, which corresponds to *smaller* lifetime-value local reservoirs that do *not* yield consecutive identical decisions (or CBL).

## Decision consistency in single-photon system and local reservoir model

As discussed in the introduction, CBL has been studied based on experimental observations in humans [9,10] and monkeys [23]. Further, Yoshihara *et al.* experimentally demonstrated a conditioned response for a fly, *drosophila*, which we consider to be another fundamental realization of CBL in living organisms [24]. In addition, an artificially constructed decision-making mechanism has been recently investigated [7,8]. This mechanism demonstrates the fact that learning behavior is directly implementable by utilizing intrinsic physical processes.

This section discusses a simple, single-photon-based, architecture design in which decision consistency in CBL behavior is produced. The notion of local reservoir is naturally introduced in the system. Figure 5(a) shows the overall system configurations, similar to the single-photon decision maker that experimentally solves the two-armed bandit problem [7]. A linearly polarized single photon impinges on a polarization beam splitter (PBS) and is detected by one of two photodetectors: PD1 and PD2. Because of the probabilistic attribute of single photons, the photon detection event occurs at a 50:50 ratio if the linear polarization is oriented by $\pi/4$ with respect to the horizontal direction. The photon detection events at PD1 (PD2) increase if single photon polarization acts toward the horizontal (vertical) direction. This can be achieved by controlling the half waveplate located in front of the PBS.

Here we assume that PD1 photon detection is directly associated with Decision 1, and that PD2 photon detection is associated with Decision 2. Let the initial single photon polarization be equal to $\pi/4$. When Decision 1 occurs, the waveplate is rotated in the horizontal direction by a certain $\Delta$. Likewise, Decision 2 leads the waveplate to be controlled towards the vertical wing with a fraction of given by $\Delta$, as schematically shown in Fig 5(b). Hence, it is more probable that the same photodetector receives single photons in subsequent measurements, which is representative of CBL behavior. In the numerical simulations, 500 consecutive decisions were made, and this sequence was repeated 100,000 times. The decision consistency is consistent with the above discussion in former sections. The decision consistency is defined as one when the decision at cycle *t* is equal to the initial decision, and defined as zero when cycle *t* is not equal to the initial decision. The amount of polarization rotation is configured by $\Delta = \pi/R$, where *R* ranges from 4–1000. When the orientation of the waveplate is configured outside the range between 0 and $\pi/2$, the subsequent decisions are terminated.

The inset in Fig 5(c) shows the average decision consistency as a function of elapsed cycles when assuming $R$ values of 5, 10, 50, and 100. With smaller $R$ values ($R=5,10$), decision consistencies exhibit large values in the initial cycles because the large amount of $\Delta$ ($=\pi/R$) drastically biases the single photon polarizations. Consequently, the orientation waveplate quickly orients vertically or horizontally, leading to the termination of the decision-making process. This yields a reduction in consistency decreases after a certain number of cycles. On the contrary, for larger $R$ values ($R=50,100$), the decision consistencies do not exhibit larger values because the smaller $\Delta$ allows the system to stay around the initial $\pi/4$ orientation.

Indeed, the $R$ values correspond to the size of the local reservoir discussed in the former sections. A larger $R$-value indicates a larger local reservoir, where the correspondence is that a small amount of $\Delta$ gives rise to *abundant fluctuations* (hence, lower degree of CBL). Conversely, smaller $R$-values indicate the presence of a smaller local reservoir, which means that a large waveplate-orientation reconfiguration immediately restrict in the subsequent decisions (hence, higher degree of CBL).

Figure 5(d) manifests such an aspect from the analysis of the active portion of a local reservoir. Here, the active portion corresponds to the degree (percentage) to which the half waveplate has rotated with respect to complete rotation to either the horizontal or vertical directions. The more the resolution increases, the more active portion decreases, which is consistent with the behavior observed in the original local reservoir model, which is shown in Fig 3(e).

## Decision consistency in human decision-making and local reservoir model

Decision making in a simple conventional cognitive brain model is summarized as schematically shown in Fig 6(a). In this figure, an experimental intervention (or stimuli) impinges on a sensory system, and then on cognitive function, the executive system, and finally motor function. The model analysis and photon system discussed previously suggest that the decision consistency of CBL observed in human decision making could be well accommodated by the local reservoir model.

In [10], Nakao *et al.* examined the behavioral analysis of CBL using 24 healthy participants. In that study, the impact of $\beta\gamma$ power (observed in the brain) on decision making was one of the highlights; in our study, we focus only on behavioral data because our primary interest is its relevance to a local reservoir. In the experiment, two professions were displayed, and each participant was asked to judge which is preferred for him or her ("Which occupation would you rather do"). Detailed protocols were given in [10].

In short, this decision-making task was repeated eight times, wherein the *decision consistency* between [the first trial and the second trial], [the second trial and the third trial], …, and [the seventh trial and the eighth trial] for each participant. Overall, the decision consistency increases as the trials progress. The red diamond shapes in Fig 6(b) depict the decision consistency of 24 participants. The notes $\times M$ means that $M$ participants exhibited the same amount of decision consistency.

The blue circular marks in Fig 6(b) show the calculated decision consistency in the local reservoir model between the first (precisely speaking, $t = 2000$) and the eighth decision ($t = 2008$), assuming a lift time of 10. The decision consistency monotonically decreases from approximately 0.7–0.5 when the size of the local reservoir spans from $N = 2$ to $N = 50$. Therefore, the empirical data depicted by the diamond-shaped marks can have a corresponding estimated size in the local reservoir, as shown in Fig 6(b). As shown in Figs 3(d) and 4(b), the internal dynamics of the local reservoir could yield a very large decision consistency, e.g., through a long lifetime; hence, the local reservoir model would accommodate the very large decision consistency value (larger than 0.7) observed in Fig. 6(b).

## Analogy between photons and the brain − Decision consistency and local reservoir model

Finally, we suggest that the entities in photon-based decision making (or photon system (denoted by P)) and human decision making (or neural system (N)) correspond with each other, as schematically shown in Fig. 7. The entities are described as follows: (i) the environment challenge is defined as incoming single photons (P) and experimental intervention (N), (ii) the sensory system (N) and the polarizer (P) share the same role, (iii) the half waveplate (P) and the cognitive function correspond to the local reservoir, (iv) the PBS (P) and the executive system (N) have similarities, and (v) the PD (P) and the motor function (N) correspond to each other. In the case of the photon system, the orientation of the polarizer and the PBS *must* be aligned prior to the experiment; otherwise, the system does not work. The critical importance of alignment is also observed in neural systems involving sensory and executive systems. In these systems, it is referred to as *sensory-map alignment* [25]. Inspired by the local reservoir model-based viewpoint, we figure out common attributes in photon and neural systems, in addition to the behavioral similarities of decision consistency in CBL; this is another benefit that the model offers.

In this study, the role of "agent," which is the entity that *reacts* to the decision, is not explicitly included, although it is one of the most important aspects of reinforcement learning [6]. Meanwhile, neural decision-making research has been extensively studied. For example, Nakao *et al.* observed different trends regarding *internally guided* decision-making problems (wherein the standard metric of decisions is individualistic), and *externally guided* decision-making problems (wherein the standard metric of decision making is shared socially) [10,26]. We consider that these issues will naturally be incorporated in local reservoir modelling because both the effect of agents and the property of given problems can be correlated with the dynamics of the local reservoir.

The present study considers decision making involving only two selections; in order for our model to be applicable to decisions involving multiple options, the proposed model must be scaled. To achieve this scaling, a hierarchical approach was proposed and experimentally demonstrated by using single photons [27]. Correspondingly, a hierarchical extension of a local reservoir is a promising principle for scalability. Additionally, it is noteworthy to mention that the local reservoir utilized in this study was based on a simple one-dimensional structure; the energy excitation paths (or arrows in Fig 2) are limited in spatially neighboring levels. Extending to a multi-dimensional local reservoir and generalized network structure is an interesting future study. In addition, it should be emphasized that hierarchical properties of local reservoirs have already been partially argued in the present study—the lifetime in the statistical modeling and the fulfilling/exciting/outgoing dynamics ($\gamma_{in}, \gamma_{up}, \gamma_{out}$) in the analytical approach reflect the properties of *the reservoir of the local reservoir*; hence, reservoir dynamics provide different decision-making tendencies, as observed in Fig 4(b). This extendibility to broader systems is one of the unique aspects of local reservoirs compared with conventional model studies [23,28]. Both theoretical and experimental endeavors are interesting future studies.

As a deeper consideration, a local reservoir could generally characterize the background mechanisms driving the cognitive abilities of living organisms and artificial systems. The relevance to the notion of "*self*" [16] and "*consciousness*" [17-19] could come into focus. Northoff *et al.* emphasizes the role of the intrinsic or spontaneous activity of the brain, e.g., its internally generated activity, rather than simply observing the apparent reactions in neurosciences [18,19,29]. The local reservoir model could serve as a mathematical framework to obtain additional insights into the computational relevance of the brain's spontaneous activity for decision making. Ultimately, the model may even be applicable in experiments that examine mental features, such as self and consciousness.

## Conclusion

In this study, we propose a local reservoir model to account for decision consistency in CBL. The model describes a phenomenon in which making a decision increases the possibility of making that same decision again in the future. The model is inspired by the viewpoint that a decision made within a visible system is affected by hidden environments, which are referred to as local reservoirs. To highlight the most simplified spatial and temporal aspects of a local reservoir, we introduce and discuss a one-dimensional model. If the size of a local reservoir is large enough, consecutive decision making will not be affected by past decisions, thus showing lower degrees of decision consistency in CBL. In contrast, with a smaller-sized local reservoir, a biased distribution is induced, which leads to high degrees of or CBL. An analytical approach to characterizing the dynamics of a local reservoir is also discussed. Furthermore, an architecture for artificially constructed CBL based on the intrinsic physical attributes of single photons is discussed, and the effect of local reservoir is numerically evaluated. Experimental observations in human decision-making tasks are also evaluated with local reservoir modelling. The architectural similarities between photon and neural systems are discussed, including the importance of alignment issues. Extension to dealing with feedback from the agent, and other decision-making problems, scalability issues are discussed as well as potential relevance and impacts to the notion of spontaneous or internally generated activity (as for instance in the case of the brain). This study creates a path toward building mathematical foundations to understand computational mechanisms by providing systematic analysis. In addition, the findings of this study suggest deeper experimental endeavors for future scientific study and application. Most importantly, by applying the local reservoir model to objects such as photons and brains, it has the potential to reveal the most basic computational mechanisms in nature.

**Acknowledgements**

This work was supported in part by the CREST program (JPMJCR17N2) from Japan Science and Technology Agency and the Core-to-Core Program A. Advanced Research Networks and the Grants-in-Aid for Scientific Research (A) (JP17H01277) from the Japan Society for the Promotion of Science. E.Y. were supported by MEXT (Ministry of Education, Culture, Sports, Science and Technology) Grant-in-Aid for the "Building of Consortia for the Development of Human Resources in Science and Technology".

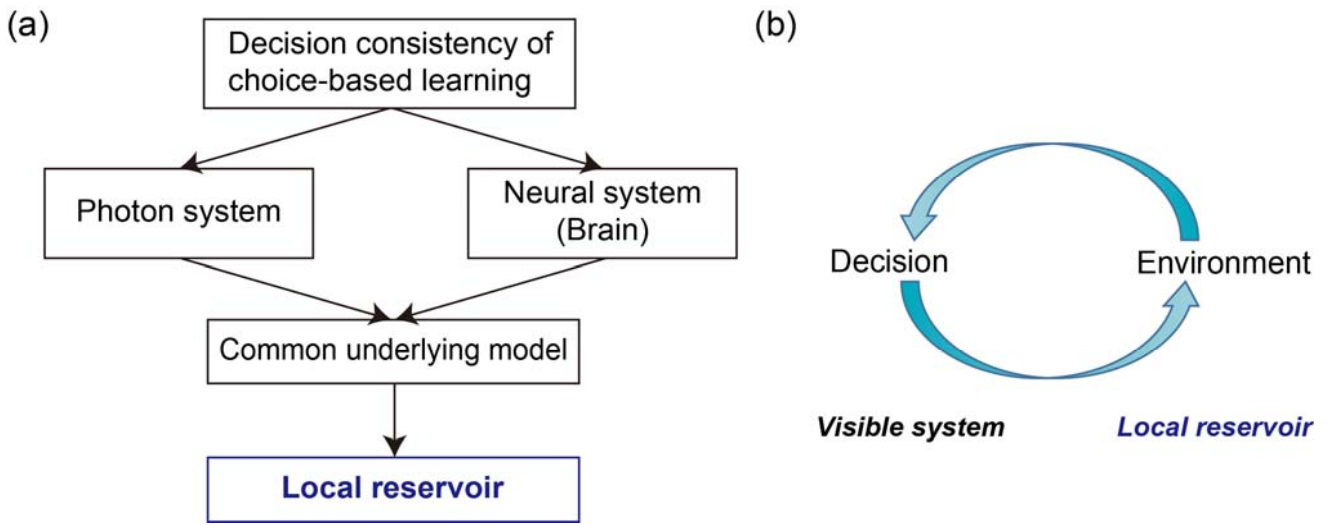

**Fig 1. Decision consistency in CBL occurs both in photon systems and neural systems; we propose the local reservoir model for a common underlying model.** (**a**) Overall approach to the subject matter. (**b**) Decision making is coupled with an environment wherein the architecture is viewed by the relation among the "visible system" and the "local reservoir."

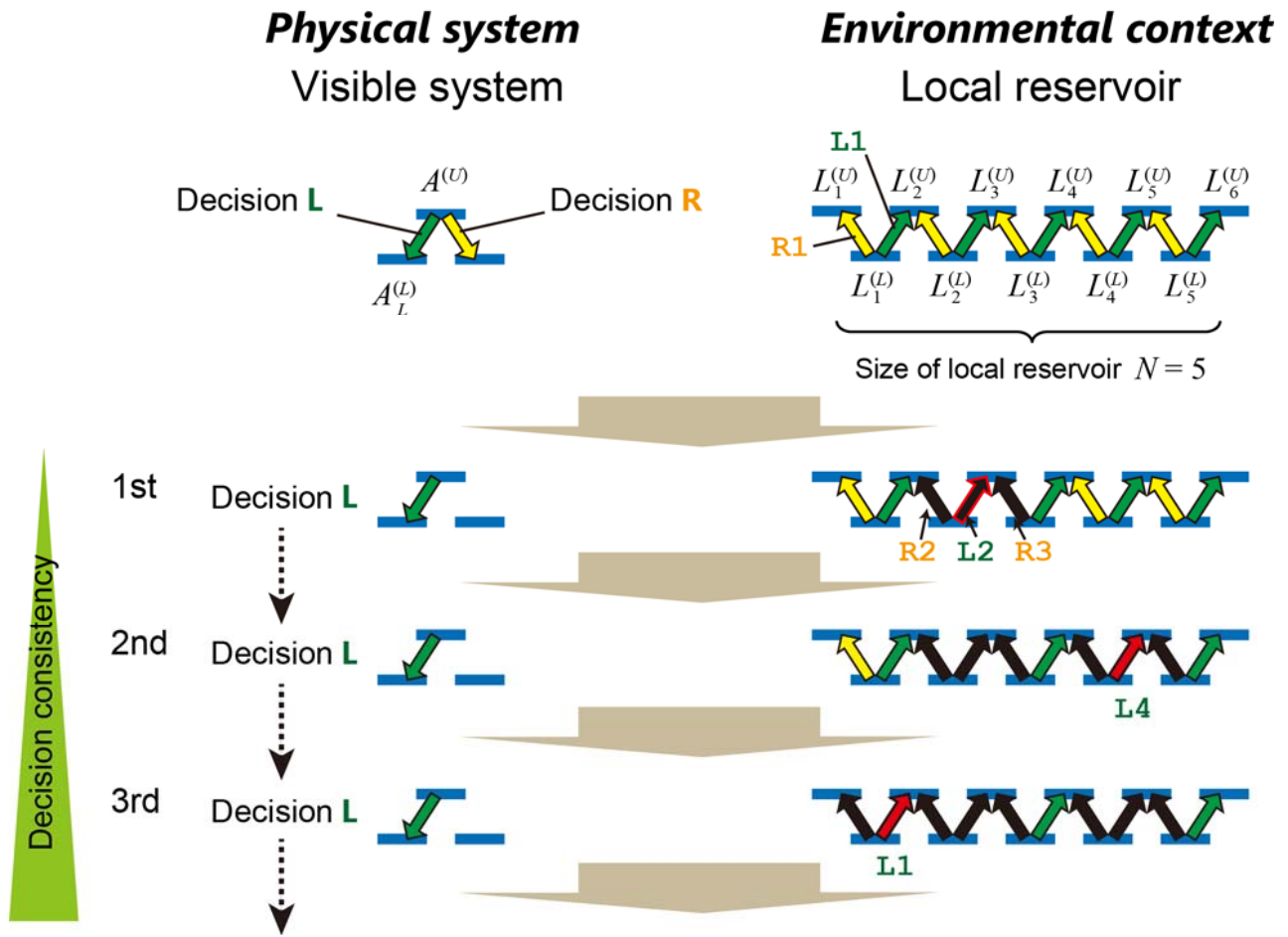

**Fig 2. Local reservoir modeling for CBL consistency.** A visible system, which acts as a physical system, is coupled with a local reservoir in which environmental context exerts influence.

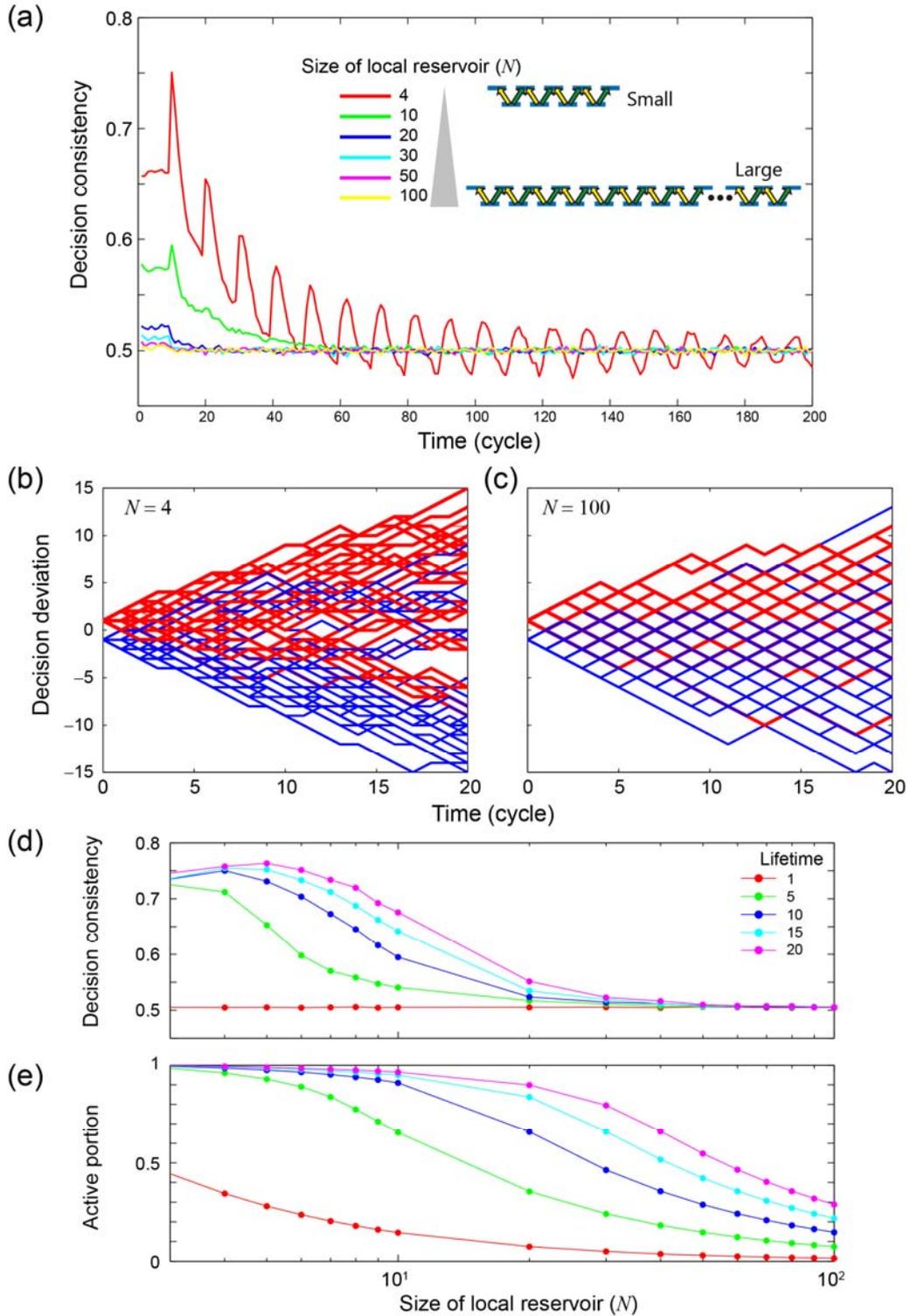

**Fig 3. CBL learning behavior via local reservoir.** (**a**) Decision consistency, which indicates the degree of CBL, exhibits higher values when the size of the local reservoir is small, whereas larger local reservoirs do not yield CBL. (**b**,**c**) Consecutive decisions are visualized in a random walk in the case of small and large local reservoirs. (**d**) The dependency to the internal dynamics of local reservoir (lifetime value). (**e**) Active portion of local reservoir as a function of the size of the local reservoir.

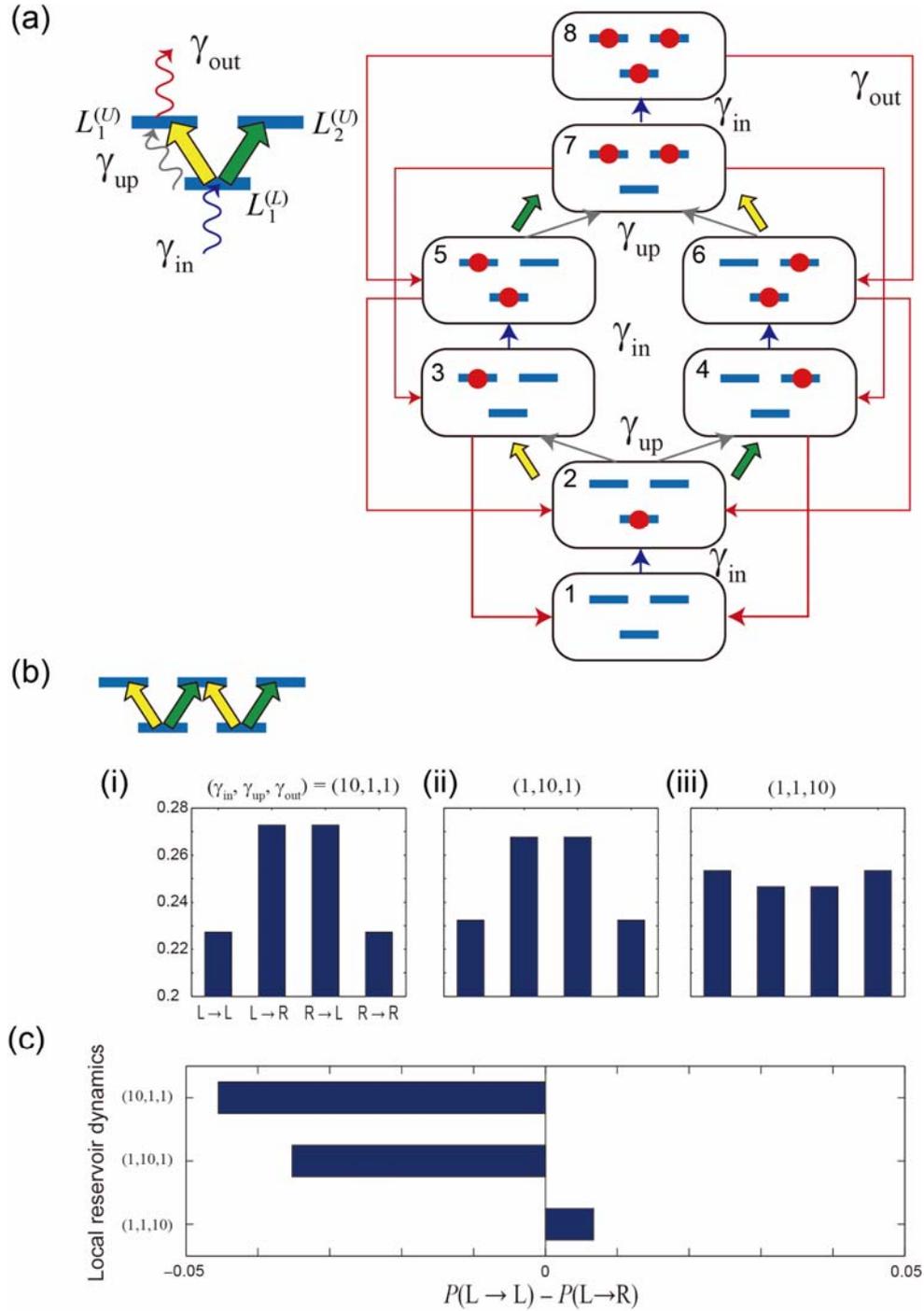

**Fig 4. Analytical modeling of local reservoir.** (a) The state transition diagram when the size of local reservoir (*N*) is 1; the number of lower energy levels is one. (b,c) When *N* = 2, the probability of making the same consecutive decision is higher than that of changing decisions when the internal dynamics of the local reservoir is slow.

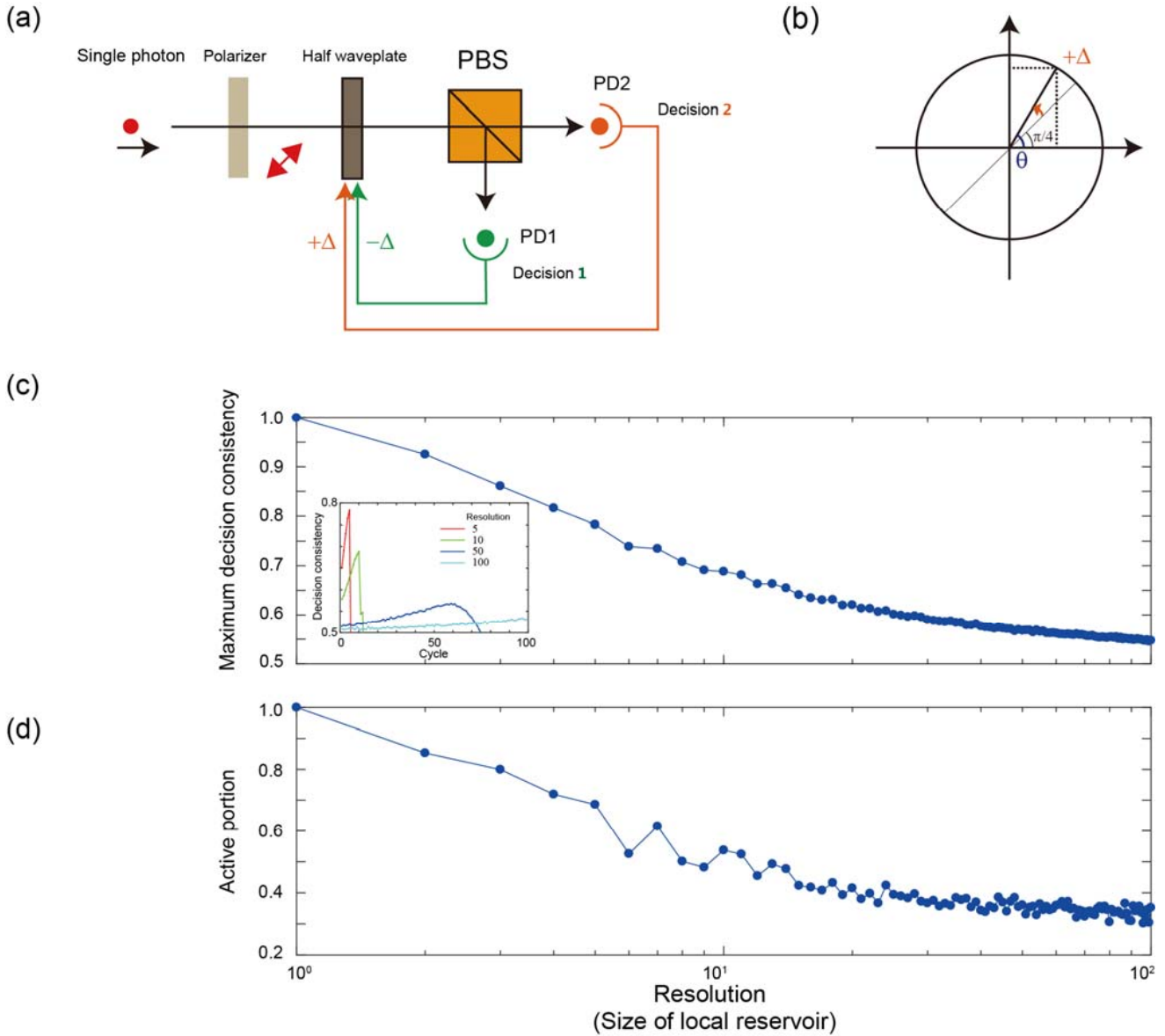

**Fig 5 Single-photon-based system exhibiting CBL** (**a**) Architecture for a decision-making system based on single photons. The angle of a linearly polarized single photon is configured by the half waveplate ($\lambda/2$). (**b**) The degree of precision (or resolution) of controlling the half waveplate. (**c**) The decision consistency, or CBL, exhibits larger values when the resolution is smaller, whereas it decreases as the resolution increases. (**d**) Active portion in the local reservoir as a function of the size of local reservoir.

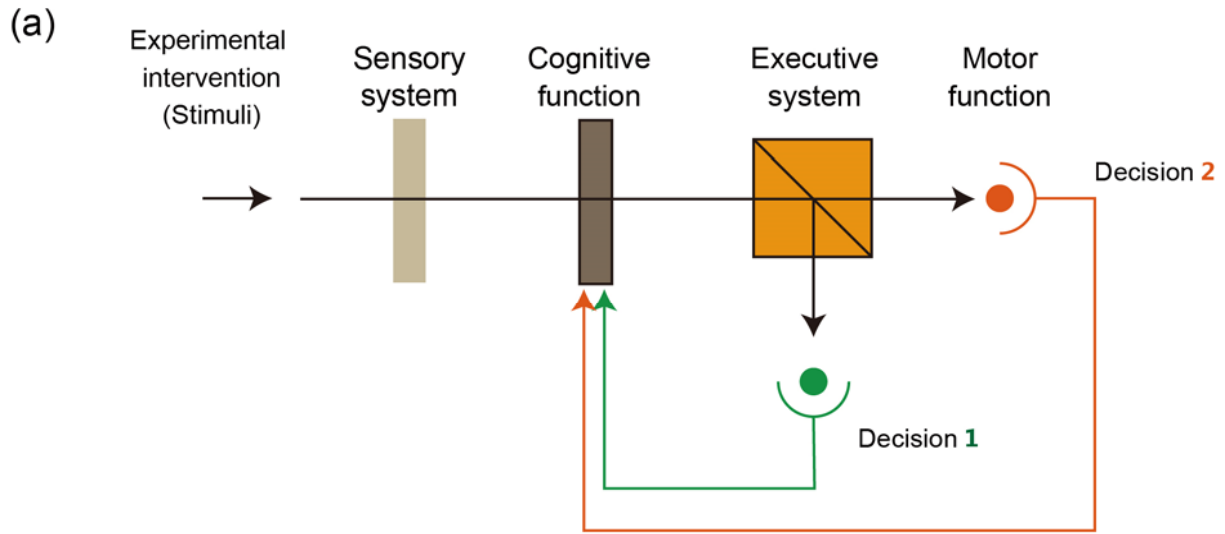

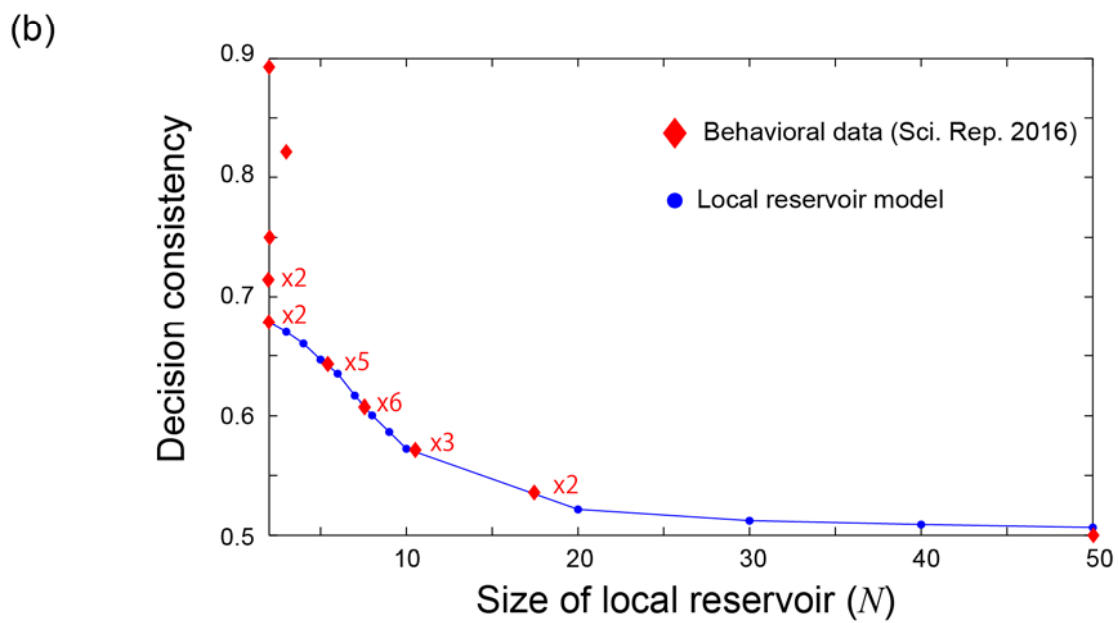

**Fig 6 Neural system exhibiting CBL** (**a**) Architecture for a decision-making system based on a cognitive model of the brain. (**b**) Decision consistency of CBL observed in human behavioral data (Occupation preference task; Nakao, *et al.* Sci. Rep. 2016 [10]) and the estimated size of local reservoir based on the model.

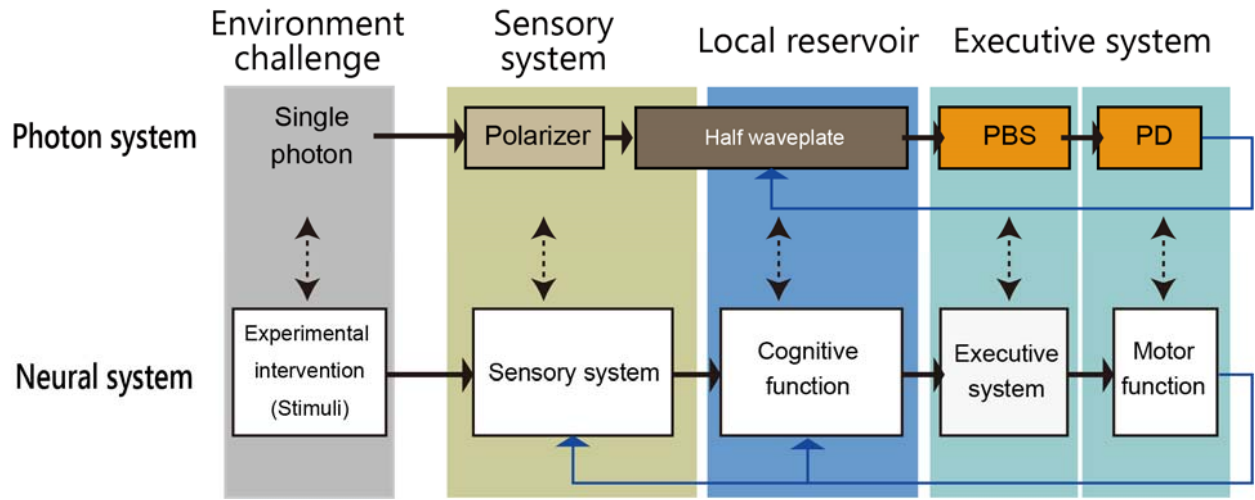

**Fig 7 Correspondence between photon system and neural system regarding CBL.**

The architecture of photon and neural systems from environmental challenge, sensory system, local reservoir, and executive system.